\documentclass[10pt,twocolumn,letterpaper]{article}

\usepackage[pagenumbers]{cvpr} % To force page numbers, e.g. for an arXiv version

\usepackage{graphicx}
\usepackage{amsmath}
\usepackage{amssymb}
\usepackage{booktabs}
\usepackage{adjustbox}
\usepackage{chngcntr}
\usepackage{multirow}
\usepackage{makecell}
\makeatletter
  \newcommand\figcaption{\def\@captype{figure}\caption}
  \newcommand\tabcaption{\def\@captype{table}\caption}
\makeatother
\usepackage{xcolor}         % colors
\usepackage{color, colortbl}
\usepackage{float}

\usepackage[pagebackref,breaklinks,colorlinks]{hyperref}

\usepackage[capitalize]{cleveref}
\crefname{section}{Sec.}{Secs.}
\Crefname{section}{Section}{Sections}
\Crefname{table}{Table}{Tables}
\crefname{table}{Tab.}{Tabs.}

\def\cvprPaperID{1201} % *** Enter the CVPR Paper ID here
\def\confName{CVPR}
\def\confYear{2023}

\begin{document}

\title{Learning 3D Representations from 2D Pre-trained Models via\\ Image-to-Point Masked Autoencoders\vspace{-0.2cm}}

\author{Renrui Zhang$^{1,2}$, Liuhui Wang$^{3}$, Yu Qiao$^{1}$, Peng Gao$^{1}$, \vspace{0.3cm} Hongsheng Li$^{2}$ \\
  $^1$Shanghai AI Laboratory \quad 
  $^2$MMLab, CUHK \quad
  $^3$Peking University\vspace{0.2cm}\\
  % $^3$The Chinese University of Hong Kong\vspace{0.2cm}\\
\texttt{\{zhangrenrui, gaopeng\}@pjlab.org.cn},\quad \texttt{hsli@ee.cuhk.edu.cn}
}
\maketitle
\begin{abstract}
\vspace{-0.1cm}
Pre-training by numerous image data has become de-facto for robust 2D representations.
In contrast, due to the expensive data acquisition and annotation, a paucity of large-scale 3D datasets severely hinders the learning for high-quality 3D features. In this paper, we propose an alternative to obtain superior 3D representations from 2D pre-trained models via \textbf{I}mage-to-\textbf{P}oint Masked Autoencoders, named as \textbf{I2P-MAE}. By self-supervised pre-training, we leverage the well learned 2D knowledge to guide 3D masked autoencoding, which reconstructs the masked point tokens with an encoder-decoder architecture. Specifically, we first utilize off-the-shelf 2D models to extract the multi-view visual features of the input point cloud, and then conduct two types of image-to-point learning schemes on top. For one, we introduce a 2D-guided masking strategy that maintains semantically important point tokens to be visible for the encoder. Compared to random masking, the network can better concentrate on significant 3D structures and recover the masked tokens from key spatial cues. For another, we enforce these visible tokens to reconstruct the corresponding multi-view 2D features after the decoder. This enables the network to effectively inherit high-level 2D semantics learned from rich image data for discriminative 3D modeling. Aided by our image-to-point pre-training, the frozen I2P-MAE, without any fine-tuning, achieves \textbf{93.4\%} accuracy for linear SVM on ModelNet40, competitive to the fully trained results of existing methods. By further fine-tuning on on ScanObjectNN's hardest split, I2P-MAE attains the state-of-the-art \textbf{90.11\%} accuracy, +3.68\% to the second-best, demonstrating superior transferable capacity. Code will be available at \url{https://github.com/ZrrSkywalker/I2P-MAE}.

\end{abstract}

%%%%%%%%% BODY TEXT
\section{Introduction}
\label{intro}

\begin{figure}[t]
  \centering
  \vspace{0.4cm}
\includegraphics[width=0.47\textwidth]{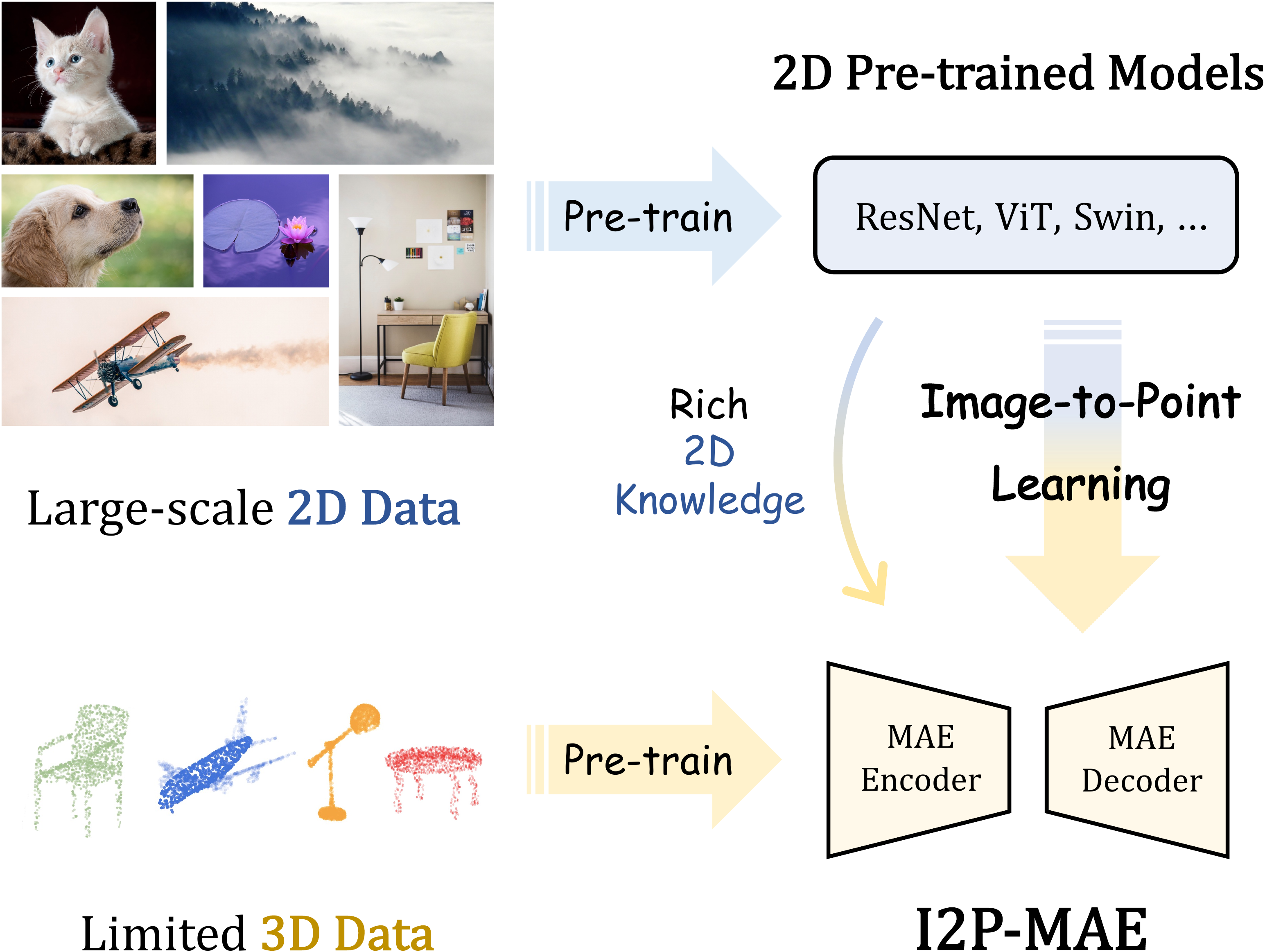}
\vspace{0.3cm}
   \caption{\textbf{Image-to-Point Masked Autoencoders.} We leverage the 2D pre-trained models to guide the MAE pre-training in 3D, which alleviates the need of large-scale 3D datasets and learns from 2D knowledge for superior 3D representations.}
    \label{fig1}
\vspace{-0.2cm}
\end{figure}

Driven by huge volumes of image data~\cite{russakovsky2015imagenet,thomee2016yfcc100m,deng2009imagenet,lin2014microsoft}, pre-training for better visual representations has gained much attention in computer vision, which benefits a variety of downstream tasks~\cite{ren2015faster,he2017mask,tian2019fcos,lin2017focal,chen2017deeplab}. Besides supervised pre-training with labels, many researches develop advanced self-supervised approaches to fully utilize raw image data via pre-text tasks, e.g., image-image contrast~\cite{caron2021emerging,he2020momentum,grill2020bootstrap,chen2021exploring,chen2020simple}, language-image contrast~\cite{clip,desai2021virtex}, and masked image modeling~\cite{mae,bao2021beit,baevski2022data2vec,xie2021simmim,gao2022convmae,huang2022green}. Given the popularity of 2D pre-trained models, it is still absent for large-scale 3D datasets in the community, attributed to the expensive data acquisition and labor-intensive annotation. The widely adopted ShapeNet~\cite{chang2015shapenet} only contains 50k point clouds of 55 object categories, far less than the 14 million ImageNet~\cite{deng2009imagenet} and 400 million image-text pairs~\cite{clip} in 2D vision. Though there have been attempts to extract self-supervisory signals for 3D pre-training~\cite{pointbert,pointcontrast,pang2022masked,jiang2022masked,liu2022masked,zhang2022point,occo,depthcontrast}, raw point clouds with sparse structural patterns cannot provide sufficient and diversified semantics compared to colorful images, which constrain the generalization capacity of pre-training. Considering the homology of images and point clouds, both of which depict certain visual characteristics of objects and are related by 2D-3D geometric mapping, we ask the question: \textit{can off-the-shelf 2D pre-trained models help 3D representation learning by transferring robust 2D knowledge into 3D domains?}

\begin{figure}[t]
  \centering
\vspace{0.05cm}
\includegraphics[width=0.44\textwidth]{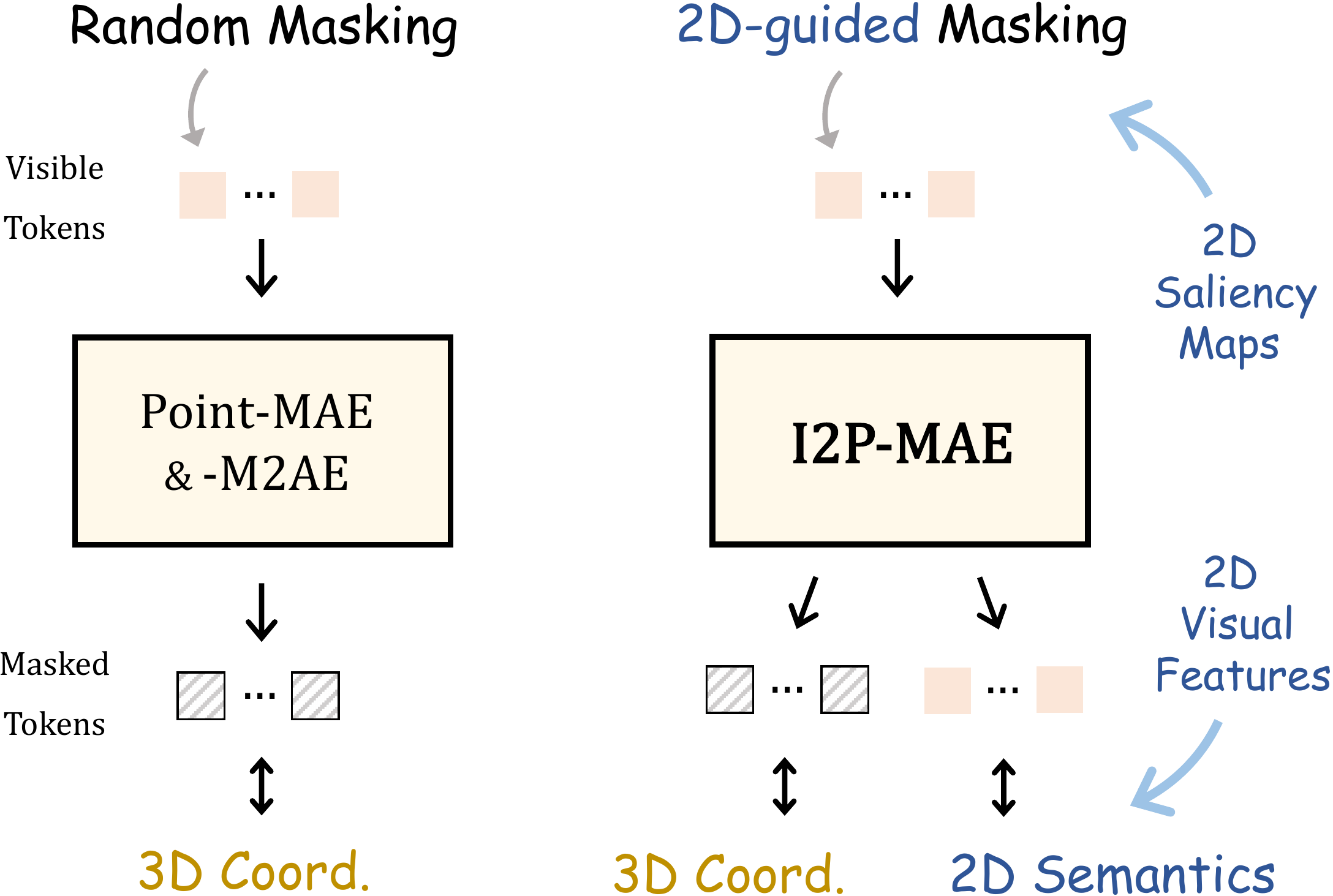}
\vspace{0.1cm}
   \caption{\textbf{Comparison of (Left) Existing Methods~\cite{pang2022masked,zhang2022point} and (Right) our I2P-MAE.} On top of the general 3D MAE architecture, I2P-MAE introduces two schemes of image-to-point learning: 2D-guided masking and 2D-semantic reconstruction.}
    \label{fig2}
% \vspace{-0.2cm}
\end{figure}

To tackle this challenge, we propose \textbf{I2P-MAE}, a Masked Autoencoding framework that conducts \textbf{I}mage-to-\textbf{P}oint knowledge transfer for self-supervised 3D point cloud pre-training. As shown in Figure~\ref{fig1}, aided by 2D semantics learned from abundant image data, our I2P-MAE produces high-quality 3D representations and exerts strong transferable capacity to downstream 3D tasks. Specifically, referring to 3D MAE models~\cite{zhang2022point,pang2022masked} in Figure~\ref{fig2} (Left), we first adopt an asymmetric encoder-decoder transformer~\cite{vit} as our fundamental architecture for 3D pre-training, which takes as input a randomly masked point cloud and reconstructs the masked points from the visible ones. Then, to acquire 2D semantics for the 3D shape, we bridge the model gap by efficiently projecting the point cloud into multi-view depth maps. This requires no time-consuming offline rendering and largely preserves 3D geometries from different perspectives. On top of that, we utilize off-the-shelf 2D models to obtain the multi-view 2D features along with 2D saliency maps of the point cloud, and respectively guide the pre-training from two aspects, as shown in Figure~\ref{fig2} (Right).

\textit{Firstly}, different from existing methods~\cite{pang2022masked,zhang2022point} to randomly sample visible tokens, we introduce a 2D-guided masking strategy that reserves point tokens with more spatial semantics to be visible for the MAE encoder. In detail, we back-project the multi-view semantic saliency maps into 3D space as a spatial saliency cloud. Each element in the saliency cloud indicates the semantic significance of the corresponding point token. Guided by such saliency cloud, the 3D network can better focus on the visible critical structures to understand the global 3D shape, and also reconstruct the masked tokens from important spatial cues. 

\textit{Secondly}, in addition to the recovering of masked point tokens, we propose to concurrently reconstruct 2D semantics from the visible point tokens after the MAE decoder. For each visible token, we respectively fetch its projected 2D representations from different views, and integrate them as the 2D-semantic learning target. By simultaneously reconstructing the masked 3D coordinates and visible 2D concepts, I2P-MAE is able to learn both low-level spatial patterns and high-level semantics pre-trained in 2D domains, contributing to superior 3D representations. 

\begin{figure}[t]
  \centering
% \vspace{0.2cm}
\includegraphics[width=0.41\textwidth]{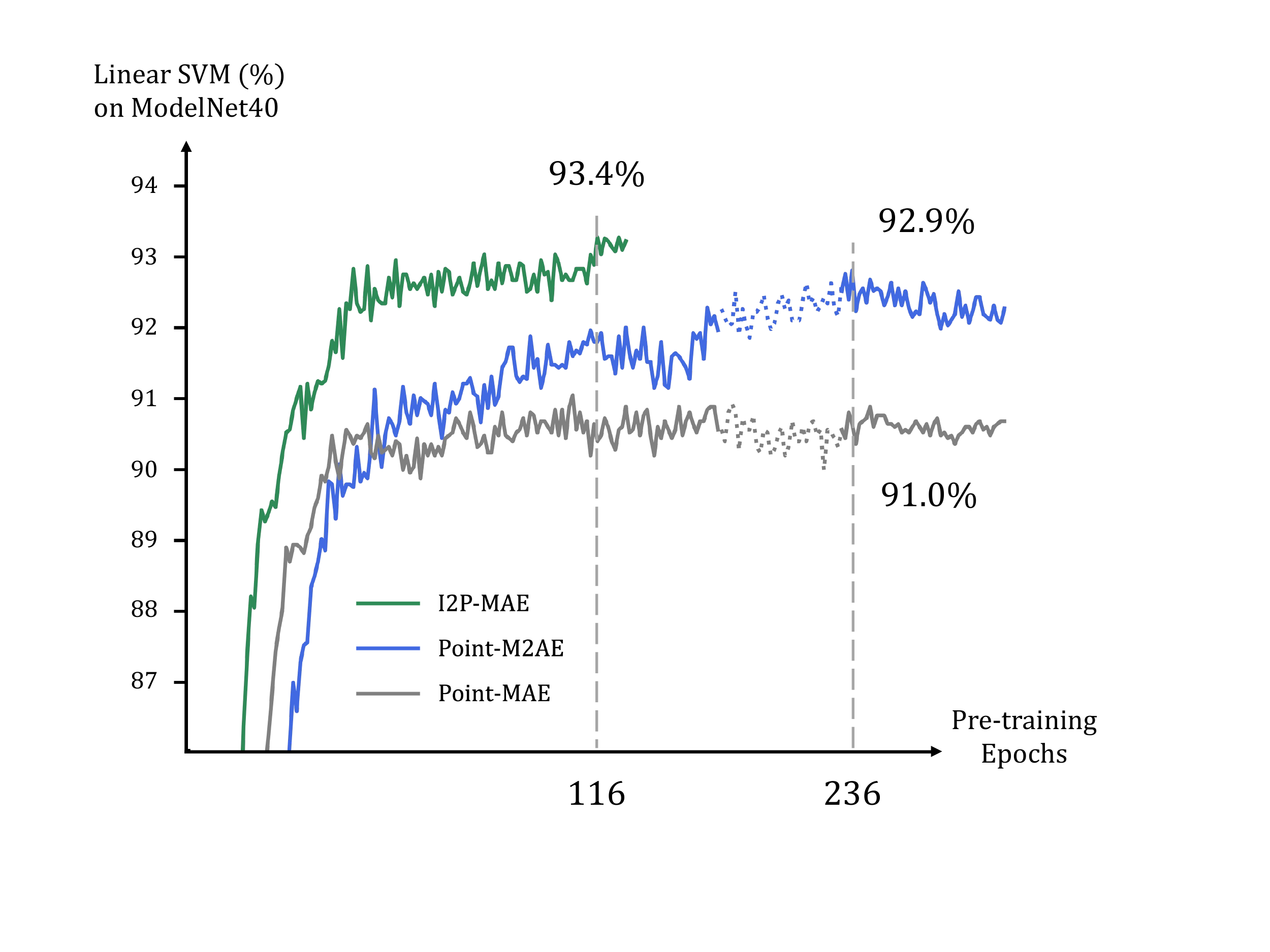}
% \vspace{-0.05cm}
   \caption{\textbf{Pre-training Epochs vs. Linear SVM Accuracy on ModelNet40~\cite{modelnet40}.} With the image-to-point learning schemes, I2P-MAE exerts superior transferable capability with much faster convergence speed than Point-MAE~\cite{pang2022masked} and Point-M2AE~\cite{zhang2022point}.}
    \label{fig3}
% \vspace{-0.2cm}
\end{figure}

With the aforementioned image-to-point guidance, our I2P-MAE significantly accelerates the convergence speed of pre-training and exhibits state-of-the-art performance on 3D downstream tasks, as shown in Figure~\ref{fig3}. Learning from 2D ViT~\cite{vit} pre-trained by CLIP~\cite{clip}, I2P-MAE, without any fine-tuning, achieves \textbf{93.4\%} classification accuracy by linear SVM on ModelNet40~\cite{modelnet40}, which has surpassed the fully fine-tuned results of Point-BERT~\cite{pointbert} and Point-MAE~\cite{jiang2022masked}. After fine-tuning, I2P-MAE further achieves \textbf{90.11\%} classification accuracy on the hardest split of ScanObjectNN~\cite{scanobjectnn}, significantly exceeding the second-best Point-M2AE~\cite{zhang2022point} by +3.68\%. The experiments fully demonstrate the effectiveness of learning from pre-trained 2D models for superior 3D representations.

Our contributions are summarized as follows:
\begin{itemize}
   \item We propose Image-to-Point Masked Autoencoders (I2P-MAE), a pre-training framework to leverage 2D pre-trained models for learning 3D representations.
   \item We introduce two strategies, 2D-guided masking and 2D-semantic reconstruction, to effectively transfer the well learned 2D knowledge into 3D domains.  
   \item Extensive experiments have been conducted to indicate the significance of our image-to-point pre-training.
\end{itemize}

\section{Related Work}

\paragraph{3D Point Cloud Pre-training.}
Supervised learning for point clouds has attained remarkable progress by delicately designed architectures~\cite{qi2017pointnet,qi2017pointnet++,guo2021pct,dgcnn} and local operators~\cite{pointmlp,curvenet,li2018pointcnn,xu2021paconv}. However, such methods learned from closed-set datasets~\cite{modelnet40,scanobjectnn} are limited to produce general 3D representations. Instead, self-supervised pre-training via unlabelled point clouds~\cite{chang2015shapenet}
have revealed promising transferable ability, which provides a good network initialization for downstream fine-tuning. Mainstream 3D self-supervised approaches adopt encoder-decoder architectures to recover the input point clouds from the transformed representations, including  point rearrangement~\cite{sauder2019self}, part occlusion~\cite{occo}, rotation~\cite{poursaeed2020self}, downsampling~\cite{li2019pu}, and codeword encoding~\cite{foldingnet}. Concurrent works also adopt contrastive pre-text tasks between 3D data pairs, such as local-global relation~\cite{rao2020global,fu2022distillation}, temporal frames~\cite{strl}, and augmented viewpoints~\cite{pointcontrast}. More recent works adopt pre-trained CLIP~\cite{clip} for zero-shot 3D recognition~\cite{zhang2021pointclip,zhu2022pointclip,guo2022calip}, or introduce masked point modeling~\cite{jiang2022masked,liu2022masked,fu2022pos} as strong 3D self-supervised learners. Therein, Point-BERT~\cite{pointbert} utilizes pre-trained tokenizers to indicate discrete point tokens, while Point-MAE~\cite{pang2022masked} and Point-M2AE~\cite{zhang2022point} apply Masked Autoencoders (MAE) to directly reconstruct the 3D coordinates of masked tokens. Our I2P-MAE also adopts MAE as the basic pre-training framework, but is guided by 2D pre-trained models via image-to-point learning schemes, which benefits 3D pre-training with diverse 2D semantics.
% \vspace{-0.2cm}

\paragraph{Masked Autoencoders.}
To achieve more efficient masked image modeling~\cite{xie2021simmim,baevski2022data2vec,zhou2021ibot,bao2021beit}, MAE~\cite{mae} is firstly proposed on 2D images with an asymmetric encoder-decoder transformer~\cite{vit}. The encoder takes as input a randomly masked image and is responsible for extracting its high-level latent representation. Then, the lightweight decoder explores informative cues from the encoded visible features, and reconstructs raw RGB pixels of the masked patches. Given its superior performance on downstream tasks, a series of follow-up works have been developed to improve MAE with customized designs: pyramid architectures with convolution stages~\cite{gao2022convmae}, window attention by grouping visible tokens~\cite{huang2022green}, high-level targets with semantic-aware sampling~\cite{hou2022milan}, and others~\cite{liu2022mixmim}. Following the spirit, Point-MAE~\cite{pang2022masked} and MAE3D~\cite{jiang2022masked} extend MAE-style pre-training on 3D point clouds, which randomly sample visible point tokens for the encoder and reconstruct masked 3D coordinates via the decoder. Point-M2AE~\cite{zhang2022point} further modifies the transformer architecture to be hierarchical for multi-scale 3D learning. Our proposed I2P-MAE aims to endow masked autoencoding on point clouds with the guidance from 2D pre-trained knowledge. By introducing the 2D-guided masking and 2D-semantic reconstruction, I2P-MAE fully releases the potential of MAE paradigm for 3D representation learning.

\paragraph{2D-to-3D Learning.}
Except for jointly training 2D-3D networks~\cite{afham2022crosspoint,liu20213d,li2022simipu,yan20222dpass}, only a few existing researches focus on 2D-to-3D learning, and can be categorized into two groups. To eliminate the modal gap, the first group either upgrades 2D pre-trained networks into 3D variants for processing point clouds (convolutional kernels inflation~\cite{zhang2019hyperspectral,xu2021image2point}, modality-agnostic transformer~\cite{qian2022pix4point}), or projects 3D point clouds into 2D images with parameter-efficient tuning (multi-view adapter~\cite{zhang2021pointclip}, point-to-pixel prompting~\cite{wang2022p2p}). Different from them, our I2P-MAE benefits from three properties. \textbf{1) Pre-training:} We learn from 2D models during the pre-training stage, and then can be flexibly adapted to various 3D scenarios by fine-tuning. However, prior works are required to utilize 2D models on evert different downstream task. \textbf{2) Self-supervision:} I2P-MAE is pre-trained by raw point clouds in a self-supervised manner and learns more general 3D representations. In contrast, prior works are supervised by labelled downstream 3D data, which might constrain the diverse 2D semantics into specific 3D domains. \textbf{3) Independence:} Prior works directly inherit 2D network architectures for 3D training, which is memory-consuming and not 3D extensible, while we train an independent 3D network and regard 2D models as semantic teachers. Another group of 2D-to-3D methods utilize paired real-world 2D-3D data of indoor~\cite{liu2021learning} or outdoor~\cite{sautier2022image} scenes, and conduct contrastive learning for knowledge transfer. Our approach also differs from them in two ways. \textbf{4) Masked Autoencoding.} I2P-MAE learns 2D semantics via the MAE-style pre-training without any contrastive loss between point-pixel pairs. \textbf{5) 3D Data Only.} We require no real-world image dataset during the pre-training and efficiently projects the 3D shape into depth maps for 2D features extraction.

\section{Method}

\begin{figure*}[t!]
  \centering
    \includegraphics[width=\textwidth]{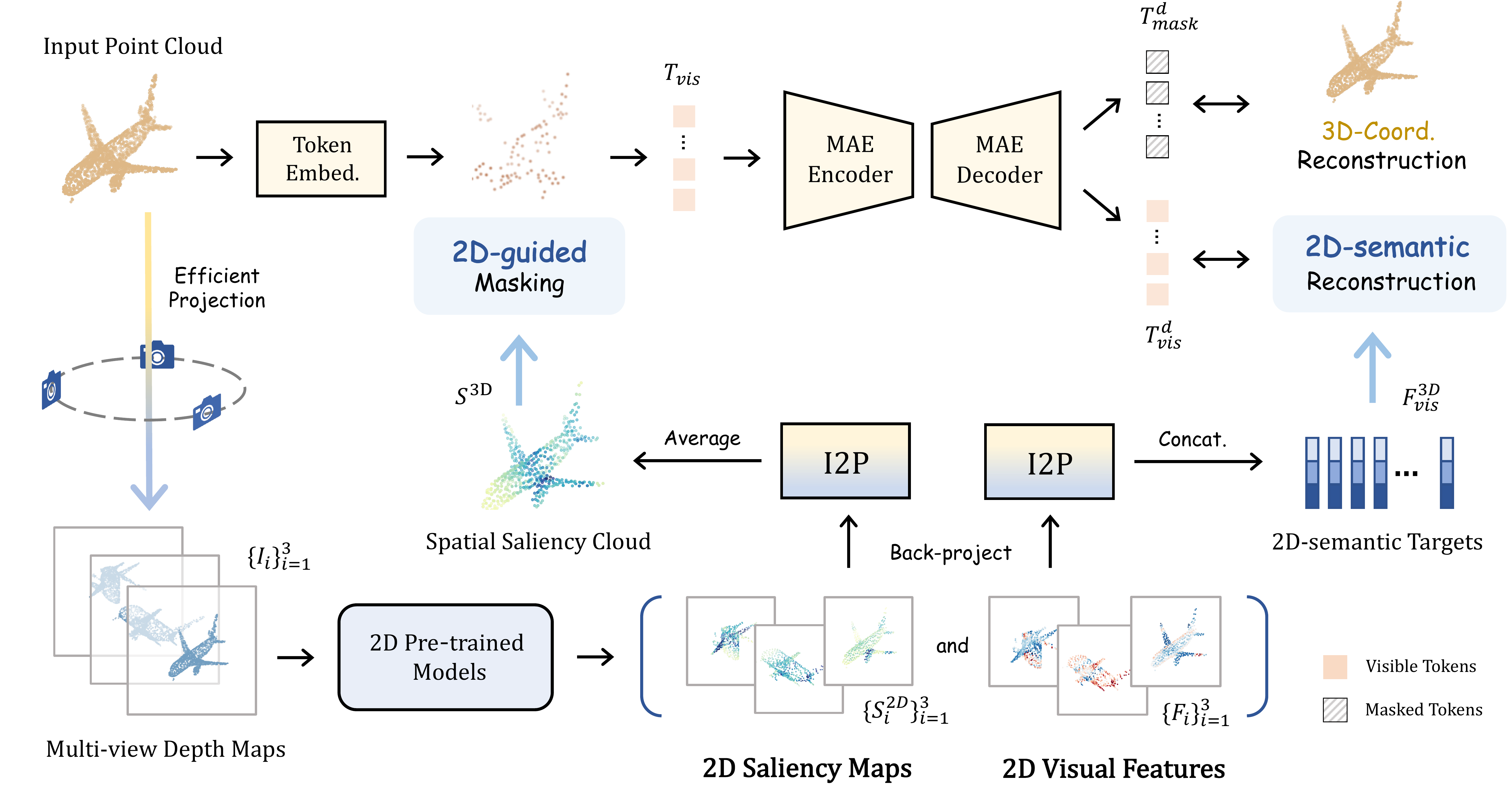}
% \vspace{0.05cm}
   \caption{\textbf{The Pipeline of I2P-MAE.} Given an input point cloud, we leverage the 2D pre-trained models to generate two guidance signals from the projected depth maps: 2D saliency maps and 2D visual features. We respectively conduct 2D-guided masking and 2D-semantic reconstruction to transfer the encoded 2D knowledge for 3D point cloud pre-training.}
    \label{fig4}
    % \vspace{-0.3cm}
\end{figure*}

The overall pipeline of I2P-MAE is shown in Figure~\ref{fig4}. In Section~\ref{s3.1}, we first introduce I2P-MAE's basic 3D architecture for point cloud masked autoencoding without 2D guidance. Then in Section~\ref{s3.2}, we show the details of utilizing 2D pre-trained models to obtain visual representations from 3D point clouds. Finally in Section~\ref{s3.3}, we present how to conduct image-to-point knowledge transfer for 3D representation learning.

\subsection{Basic 3D Architecture}
\label{s3.1}

As our basic framework for image-to-point learning, I2P-MAE refers to existing works~\cite{zhang2022point,pang2022masked} to conduct 3D point cloud masked autoencoding, which consists of a token embedding module, an encoder-decoder transformer, and a head for reconstructing masked 3D coordinates. 
% \vspace{-0.2cm}

\paragraph{Token Embedding.} 
Given a raw point cloud $P \in \mathbb{R}^{N\times 3}$, we first adopt Furthest Point Sampling (FPS) to downsample the point number from $N$ to $M$, denoted as $P^T \in \mathbb{R}^{M\times 3}$. Then, we utilize $k$ Nearest-Neighbour ($k$-NN) to search the $k$ neighbors for each downsampled point, and aggregate their features via a mini-PointNet~\cite{qi2017pointnet} to obtain $M$ point tokens. In this way, each token can represent a local spatial region and interact long-range features with others in the follow-up transformer. We formulate them as $T \in \mathbb{R}^{M\times C}$, where $C$ denotes the feature dimension.
% \vspace{-0.2cm}

\paragraph{Encoder-Decoder Transformer.}
To build the pre-text learning targets, we mask the point tokens with a high ratio, e.g., 80\%, and only feed the visible ones, $T_{\rm vis} \in \mathbb{R}^{M_{\rm vis}\times C}$, into the transformer encoder, where $M_{\rm vis}$ denotes the visible token number. Each encoder block contains a self-attention layer and is pre-trained to understand the global 3D shape from the remaining visible parts. After encoding, we concatenate the encoded $T^e_{\rm vis}$ with a set of shared learnable masked tokens $T_{\rm mask} \in \mathbb{R}^{M_{\rm mask}\times C}$, and input them into a lightweight decoder, where $M_{mask}$ denotes the masked token number and $M = M_{\rm mask} + M_{\rm vis}$. In the transformer decoder, the masked tokens learn to capture informative spatial cues from visible ones and reconstruct the masked 3D coordinates. Importantly, we follow Point-M2AE~\cite{zhang2022point} to modify our transformer to be a hierarchical architecture for encoding the multi-scale representations.

\paragraph{3D-coordinate Reconstruction.}
On top of the decoded point tokens $\{T^d_{\rm vis}, T^d_{\rm mask}\}$, we utilize $T^d_{\rm mask}$ to reconstruct 3D coordinates of the masked tokens along with their $k$ neighboring points. A reconstruction head of a single linear projection layer is adopted to predict $P_{\rm mask} \in \mathbb{R}^{M_{\rm mask}\times k\times 3}$, the ground-truth 3D coordinates of the masked points. Then, we compute the loss by Chamfer Distance~\cite{chamfer} and formulate it as
\begin{align}
    \mathcal{L}_{\rm 3D} = \frac{1}{M_{\rm mask}k}\text{Chamfer}\Big(\operatorname{H_{3D}}(T^d_{\rm mask}), {P}_{\rm mask}\Big),
\end{align}
where $\operatorname{H_{3D}}(\cdot)$ denotes the head to reconstruct the masked 3D coordinates.

\subsection{2D Pre-trained Representations}
\label{s3.2}
We can leverage 2D models of different architectures (ResNet~\cite{resnet}, ViT~\cite{vit}) and various pre-training approaches (supervised~\cite{resnet,liu2021swin} and self-supervised~\cite{clip,caron2021emerging} ones) to assist the 3D representation learning. To align the input modality for 2D models, we project the input point cloud onto multiple image planes to create depth maps, and then encode them into multi-view 2D representations.

\paragraph{Efficient Projection.}
To ensure the efficiency of pre-training, we simply project the input point cloud $P$ from three orthogonal views respectively along the $x, y, z$ axes. For every point, we directly omit each of its three coordinates and round down the other two to obtain the 2D location on the corresponding map. The projected pixel value is set as the omitted coordinate to reflect relative depth relations of points, which is then repeated by three times to imitate the three-channel RGB. The projection of I2P-MAE is highly time-efficient and involves no offline rendering~\cite{phong1975illumination,su2018adeeper}, projective transformation~\cite{simpleview,zhang2021pointclip}, or learnble prompting~\cite{wang2022p2p}. We denote the projected multi-view depth maps of $P$ as $\{I_i\}_{i=1}^3$.

\paragraph{2D Visual Features.}
We then utilize a 2D pre-trained model, e.g., a pre-trained ResNet or ViT, to extract multi-view features of the point cloud with $C$ channels, formulated as $\{F^{\rm 2D}_i\}_{i=1}^3$, where $F_i \in \mathbb{R}^{H\times W\times C}$ and $H, W$ denote the feature map size. Such 2D features contain sufficient high-level semantics learned from large-scale image data. The geometric information loss during projection can also be alleviated by encoding from different views. 

\paragraph{2D Saliency Maps.}
In addition to 2D features, we also acquire a semantic saliency map for each view via the 2D pre-trained model. The one-channel saliency maps indicate the semantic importance of different image regions, which we denote as $\{S^{\rm 2D}_i\}_{i=1}^3$, where $S_i \in \mathbb{R}^{H\times W\times 1}$. For ResNet, we obtain the maps by using pixel-wise max pooling on $\{F_i^{\rm 2D}\}_{i=1}^3$ to reduce the feature dimension to one. For ViT, we adopt attention maps of the class token at the last transformer layer, since the attention weights to the class token reveal how much the features contribute to the final classification.
% \vspace{0.1cm}

\subsection{Image-to-Point Learning Schemes}
\label{s3.3}

On top of the 2D pre-trained representations for the point cloud, I2P-MAE's pre-training is guided by two image-to-point learning designs: 2D-guided masking before the encoder, and 2D-semantic reconstruction after the decoder.

\begin{figure}[t]
  \centering
% \vspace{0.2cm}
\includegraphics[width=0.45\textwidth]{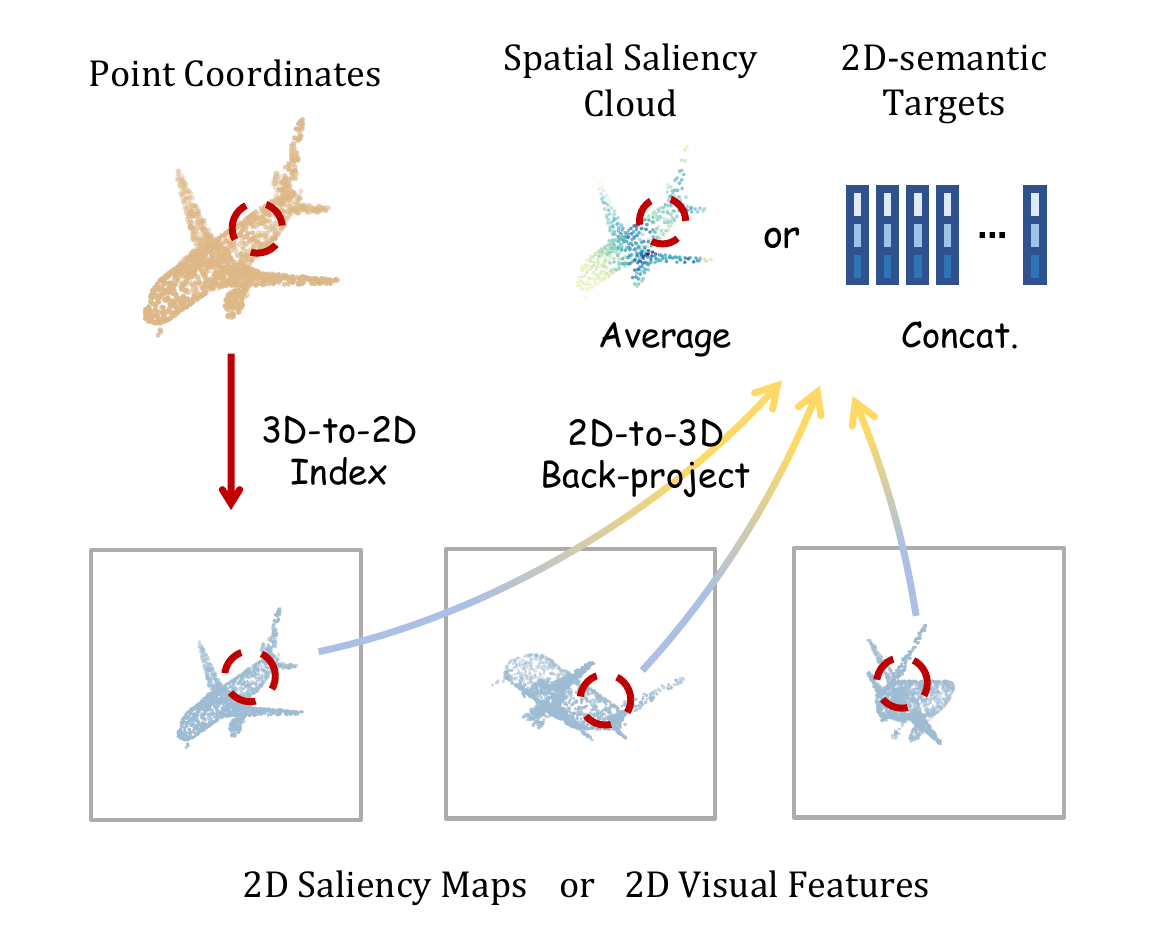}
% \vspace{0.05cm}
   \caption{\textbf{Image-to-Point Operation (I2P).} Indexed by 3D point coordinates, the corresponding multi-view 2D representations are back-projected into 3D space for aggregation.}
    \label{fig5}
% \vspace{-0.2cm}
\end{figure}

\paragraph{2D-guided Masking.}
The conventional masking strategy samples masked tokens randomly following a uniform distribution, which might prevent the encoder from `seeing' important spatial characteristics and disturb the decoder by nonsignificant structures. Therefore, we leverage the 2D semantic saliency maps to guide the masking of point tokens, which samples more semantically significant 3D parts for encoding. Specifically, indexed by the coordinates of point tokens $P^T \in \mathbb{R}^{M\times 3}$, we back-project the multi-view saliency maps $\{S^{\rm 2D}_i\}_{i=1}^3$ into 3D space and aggregate them as a 3D semantic saliency cloud $S^{\rm 3D} \in \mathbb{R}^{M\times 1}$. For each point in $S^{\rm 3D}$, we assign the semantic score by averaging the corresponding 2D values from multi-view saliency maps, formulated as
\begin{align}
    S^{\rm 3D} = \operatorname{Softmax}\Big(\frac{1}{3}\sum_{i=1}^3\operatorname{I2P}(S^{\rm 2D}_i, P^T)\Big),
\end{align}
where $\operatorname{I2P}(\cdot)$ denotes the 2D-to-3D back-projection operation in Figure~\ref{fig5}. We apply a softmax function to normalize the $M$ points within $S^{\rm 3D}$, and regard each element's magnitude as the visible probability for the corresponding point token. With this 2D semantic prior, the random masking becomes a nonuniform sampling with different probabilities for different tokens, where the tokens covering more critical 3D structures are more likely to be preserved. This boosts the representation learning of the encoder by more focusing on significant 3D geometries, and provides the masked tokens with more informative cues at the decoder for better reconstruction.
% \vspace{-0.2cm}

\paragraph{2D-semantic Reconstruction.}
The 3D-coordinate reconstruction of masked point tokens enables the network to explore low-level 3D patterns. On top of that, we further enforce the visible point tokens, $T^d_{\rm vis}$, to reconstruct the extracted 2D semantics from different views, which effectively transfers the 2D pre-trained knowledge into 3D pre-training. To obtain the 2D-semantic targets, we utilize the coordinates of visible point tokens $P^T_{\rm vis} \in \mathbb{R}^{M_{\rm vis}\times 3}$ as indices to aggregate corresponding 2D features from $\{F^{\rm 2D}_i\}_{i=1}^3$ by channel-wise concatenation, formulated as
\begin{align}
    F_{\rm vis}^{\rm 3D} = \operatorname{Concat}\Big\{\operatorname{I2P}(F^{\rm 2D}_i, P^T_{\rm vis})\Big\}_{i=1}^3,
\end{align}
where $F_{\rm vis}^{\rm 3D} \in \mathbb{R}^{M_{\rm vis}\times 3C}$. As the multi-view depth maps depict a 3D shape from different perspectives, the concatenation between multi-view 2D features can better integrate the rich semantics inherited from 2D pre-trained models. Then, we also adopt a reconstruction head of a single linear projection layer for $T^d_{\rm vis}$, and compute the $l_2$ loss as
\begin{align}
    \mathcal{L}_{\rm 2D} = \frac{1}{M_{\rm vis}}\Big(\operatorname{H_{2D}}(T^d_{\rm vis}) - F^{\rm 3D}_{\rm vis}\Big)^2,
\end{align}
where $\operatorname{H_{2D}}(\cdot)$ denotes the head to reconstruct visible 2D semantics, parallel to $\operatorname{H_{3D}}(\cdot)$ for masked 3D coordinates. The final pre-training loss of our I2P-MAE is then formulated as $\mathcal{L}_{\rm I2P} = \mathcal{L}_{\rm 3D} + \mathcal{L}_{\rm 2D}$. With such image-to-point feature learning, I2P-MAE not only encodes low-level spatial variations from 3D coordinates, but also explores high-level semantic knowledge from 2D representations. As the 2D-guided masking has preserved visible tokens with more spatial significance, the 2D-semantic reconstruction upon them further benefits I2P-MAE to learn more discriminative 3D representations.

\begin{table}[t!]
% \vspace{0.2cm}
\small
\centering
\begin{adjustbox}{width=0.93\linewidth}
	\begin{tabular}{lcc}
	\toprule
		\makecell*[l]{Method} &ModelNet40 &\ \ \ OBJ-BG\ \ \ \\
        \cmidrule(lr){1-1} \cmidrule(lr){2-2} \cmidrule(lr){3-3}
        3D-GAN~\cite{3dgan}  &83.3 &- \\
        Latent-GAN~\cite{latentgan}   & 85.7 &- \\
        % MRTNet~\cite{mrtnet}  & 86.4 \\
        SO-Net~\cite{sonet}   &87.3 &- \\
        FoldingNet~\cite{foldingnet}  &88.4 &-\\
        % MAP-VAE~\cite{mapvae}  & 88.4 &-\\
        % 3D-PointCapsNet~\cite{}  & 88.9 \\
		VIP-GAN~\cite{vipgan}   &90.2 &-\\
		\cmidrule(lr){1-3}
		DGCNN + Jigsaw~\cite{jiasaw}  &90.6 &59.5 \\
		DGCNN + OcCo~\cite{occo}  &90.7 &78.3 \\
        DGCNN + STRL~\cite{strl}  &90.9 &77.9\\
        DGCNN + CrossPoint~\cite{afham2022crosspoint}  &91.2 &81.7\\
        \cmidrule(lr){1-3}
        Transformer + OcCo~\cite{pointbert} &89.6 &- \\
        Point-BERT~\cite{pointbert}  &87.4 &-\\
        Point-MAE~\cite{pang2022masked}  &91.0 &77.7\\
        Point-M2AE~\cite{zhang2022point}\vspace{0.1cm} &\underline{92.9} &\underline{84.1}\\
		\rowcolor{pink!12}\textbf{I2P-MAE}\vspace{0.1cm} &\textbf{93.4} &\textbf{87.1} 
		\\
		\textit{Improvement}  &\textcolor{blue}{+0.5} &\textcolor{blue}{+3.0}\\
	  \bottomrule
	\end{tabular}
\end{adjustbox}
\caption{\textbf{Linear SVM Classification on ModelNet40~\cite{modelnet40} and ScanObjectNN~\cite{scanobjectnn}}. We compare the accuracy (\%) of existing self-supervised methods, and the second-best one is underlined. We adopt the OBJ-BG split for ScanObjectNN following~\cite{afham2022crosspoint}.}
\label{t1}
% \vspace{-0.1cm}
\end{table}

\section{Experiments}

In Section~\ref{s4.1}, we first introduce our pre-training settings and the linear SVM classification performance without fine-tuning. Then in Section~\ref{s4.2}, we present the results by fully fine-tuning I2P-MAE on various 3D downstream tasks. Finally in Section~\ref{s4.3}, we conduct ablation studies to investigate the characteristics of I2P-MAE.

\subsection{Image-to-Point Pre-training}
\label{s4.1}

\paragraph{Settings.}
We adopt the popular ShapeNet~\cite{chang2015shapenet} for self-supervised 3D pre-training, which contains 57,448 synthetic point clouds with 55 object categories. For fair comparison, we follow the same MAE transformer architecture as Point-M2AE~\cite{zhang2022point}: a 3-stage encoder with 5 blocks per stage, a 2-stage decoder with 1 block per stage, 2,048 input point number ($N$), 512 downsampled number ($M$), 16 nearest neighbors ($k$), 384 feature channels ($C$), and the mask ratio of 80\%. For off-the-shelf 2D models, we utilize ViT-Base~\cite{vit} pre-trained by CLIP~\cite{clip} as default and keep its weights frozen during 3D pre-training. We project the point cloud into three $224\times 224$ depth maps, and obtain the 2D feature size $H\times W$ of $14\times 14$. I2P-MAE is pre-trained for 300 epochs with a batch size 64 and learning rate $10^{-3}$. We adopt AdamW~\cite{kingma2014adam} optimizer with a weight decay $5 \times 10^{-2}$ and the cosine scheduler with 10-epoch warm-up.

\begin{table}[t!]
% \vspace{0.2cm}
\small
\centering
\begin{adjustbox}{width=\linewidth}
	\begin{tabular}{lccc}
	\toprule
		\makecell*[l]{Method} &OBJ-BG &OBJ-ONLY &PB-T50-RS\\
		\cmidrule(lr){1-1} \cmidrule(lr){2-2} \cmidrule(lr){3-3} \cmidrule(lr){4-4}
% 		\multicolumn{4}{c}{Train from Scratch} \\
% 		\cmidrule(lr){1-4}
	    PointNet~\cite{qi2017pointnet}  &73.3 &79.2 &68.0\\
	    SpiderCNN~\cite{xu2018spidercnn}  &77.1 &79.5 &73.7\\
	    PointNet++~\cite{qi2017pointnet++} &82.3 &84.3 &77.9\\
	    DGCNN~\cite{dgcnn} &82.8 &86.2 &78.1\\
	    PointCNN~\cite{li2018pointcnn} &86.1 &85.5 &78.5\\
	   % BGA-DGCNN~\cite{} &- &- &79.7 \\
	   % BGA-PN++~\cite{} &- &- &80.2 \\
	   % DRNet~\cite{drnet} &- &- &80.3\\
	   % GBNet~\cite{gbnet} &- &-  &80.5\\
	    SimpleView~\cite{simpleview} &- &-  &80.5\\
	    MVTN~\cite{hamdi2021mvtn} &- &-  &82.8\\
	    PointMLP~\cite{pointmlp} &- &-  &85.2\\
	   % \midrule
	   \cmidrule(lr){1-4}
	   % \multicolumn{4}{c}{Fine-tune after Pre-training} \\
% 		\cmidrule(lr){1-4}
	    Transformer~\cite{pointbert} &79.86 &80.55 &77.24  \\
	   % {[P]} OcCo~\cite{occo} &84.85 &85.54 &78.79\vspace{0.05cm}\\
	    {[P]} Point-BERT~\cite{pointbert} &87.43 &88.12 &83.07\\
	    {[P]} MaskPoint~\cite{liu2022masked} &89.30 &88.10 &84.30\\
	    {[P]} Point-MAE~\cite{pang2022masked} &90.02 &88.29 &85.18\\
	    {[P]} MAE3D~\cite{jiang2022masked} &- &- &86.20\\
	    {[P]} Point-M2AE~\cite{zhang2022point}\vspace{0.1cm} &\underline{91.22} &\underline{88.81} &\underline{86.43}\\
	    \rowcolor{pink!12} \textbf{[P] I2P-MAE}\vspace{0.1cm} &\textbf{94.15} &\textbf{91.57}  &\textbf{90.11}\\
	     \textit{Improvement} &\textcolor{blue}{+2.93} &\textcolor{blue}{+2.76} &\textcolor{blue}{+3.68} \\
	  \bottomrule
	\end{tabular}
\end{adjustbox}
\caption{\textbf{Real-world 3D Classification on ScanObjectNN~\cite{scanobjectnn}}. We report the accuracy (\%) on the official three splits of ScanObjectNN. [P] denotes to fine-tune the models after self-supervised pre-training.}
\label{t2}
% \vspace{-0.1cm}
\end{table}

\paragraph{Linear SVM.}\vspace{-0.2cm}
To evaluate the transfer capacity, we directly utilize the features extracted by I2P-MAE's encoder for linear SVM on the synthetic ModelNet40~\cite{modelnet40} and real-world ScanObjectNN~\cite{scanobjectnn} without any fine-tuning or voting. As shown in Table~\ref{t1}, for 3D shape classification in both domains, I2P-MAE shows superior performance and exceeds the second-best respectively by +0.5\% and +3.0\% accuracy. Our SVM results (93.4\%, 87.1\%) can even surpass some existing methods after full downstream training in Table~\ref{t2} and~\ref{t3}, e.g., PointCNN~\cite{li2018pointcnn} (92.2\%, 86.1\%), Transformer~\cite{pointbert} (91.4\%, 79.86\%), and Point-BERT~\cite{pointbert} (92.7\%, 87.43\%). In addition, guided by the 2D pre-trained models, I2P-MAE exhibits much faster pre-training convergence than Point-MAE~\cite{pang2022masked} and Point-M2AE in Figure~\ref{fig3}.
Therefore, the SVM performance of I2P-MAE demonstrates its learned high-quality 3D representations and the significance of our image-to-point learning schemes.

\subsection{Downstream Tasks}
\label{s4.2}
After pre-training, I2P-MAE is fine-tuned for real-world and synthetic 3D classification, and part segmentation. Except ModelNet40~\cite{modelnet40}, we do not use the voting strategy~\cite{rscnn} for evaluation.
\vspace{-0.1cm}

\paragraph{Real-world 3D Classification.}
The challenging ScanObjectNN~\cite{scanobjectnn} consists of 11,416 training and 2,882 test 3D shapes, which are scanned from the real-world scenes and thus include backgrounds with noises. As shown in Table~\ref{t2}, our I2P-MAE exerts great advantages over other self-supervised methods, surpassing the second-best by +2.93\%, +2.76\%, and +3.68\% respectively for the three splits. This is also the first model reaching 90\% accuracy on the hardest PB-T50-RS spilt.
As the pre-training point clouds~\cite{chang2015shapenet} are synthetic 3D shapes with a large domain gap with the real-world ScanobjectNN, the results well indicate the universality of I2P-MAE inherited from the 2D pre-trained models.

\begin{table}[t!]
% \vspace{0.2cm}
\small
\centering
\begin{adjustbox}{width=0.9\linewidth}
	\begin{tabular}{lccc}
	\toprule
		\makecell*[l]{Method} &Points &No Voting &Voting\\
		\cmidrule(lr){1-1} \cmidrule(lr){2-2} \cmidrule(lr){3-3} \cmidrule(lr){4-4}
% 		\multicolumn{4}{c}{Train from Scratch} \\
% 		\cmidrule(lr){1-4}
	     PointNet~\cite{qi2017pointnet}  &1k &89.2 &- \\ 
        PointNet++~\cite{qi2017pointnet++}  &1k &90.7 &- \\
        PointCNN~\cite{li2018pointcnn}  &1k &92.2 &- \\
        % {[P]} SO-Net~\cite{sonet}  &5k &92.5 &-\\
        % DensePoint~\cite{}  &1k &92.8 \\
        DGCNN~\cite{dgcnn}  &1k &92.9 &-\\
        RSCNN~\cite{rscnn}  &1k &92.9 &93.6 \\
        % KPConv~\cite{thomas2019kpconv}  &6k &92.9 \\
        PCT~\cite{guo2021pct}  &1k &93.2 &-\\
        % SimpleView~\cite{}
        Point Transformer~\cite{pointtransformer}  &- &93.7 \\
        % CurveNet~\cite{}  &1k &93.8 \\
        \cmidrule(lr){1-4}
        % \multicolumn{3}{c}{Fine-tune after Pre-training} \\
% 		\cmidrule(lr){1-3}
        % Transformer~\cite{pointbert}  &1k &91.4\\
        % {[P] Transformer + OcCo}~\cite{pointbert}  &1k &92.1\vspace{0.05cm}\\
        {[P] Point-BERT}~\cite{pointbert}  &1k &92.7 &93.2\\
        \color{gray}{\ \ \ \ \ \ Point-BERT}  &\color{gray}{4k} &\color{gray}{92.9} &\color{gray}{93.4}\\
        \color{gray}{\ \ \ \ \ \ Point-BERT}  &\color{gray}{8k} &\color{gray}{93.2} &\color{gray}{93.8}\\
        {[P] MAE3D}~\cite{jiang2022masked}  &1k &- &93.4\\
        {[P] MaskPoint}~\cite{liu2022masked}  &1k &- &93.8\\
        {[P] Point-MAE~\cite{pang2022masked}}  &1k &93.2 &93.8\\
        \color{gray}{\ \ \ \ \ \ Point-MAE}  &\color{gray}{8k} &\color{gray}{-} &\color{gray}{94.0}\\
        {[P] Point-M2AE~\cite{zhang2022point}}\vspace{0.05cm}  &1k &93.4 &94.0\\
       {[P] I2P-MAE-svm}\vspace{0.05cm} &{1k} &{93.4} &{-}\\
        % \rowcolor{blue!4}{\ \ \ \ \ \ I2P-MAE-mlp} &1k &93.6 &-\\
        \rowcolor{pink!12}\textbf{\ \ \ \ \ \ I2P-MAE}\vspace{0.1cm} &\textbf{1k} &\textbf{93.7} &\textbf{94.1}\\
	  \bottomrule
	\end{tabular}
\end{adjustbox}
\caption{\textbf{Synthetic 3D Classification on ModelNet40~\cite{modelnet40}}. We report the accuracy (\%) before and after the voting~\cite{rscnn}. [P] denotes to fine-tune the models after self-supervised pre-training.}
\label{t3}
% \vspace{-0.1cm}
\end{table}

\paragraph{Synthetic 3D Classification.}\vspace{-0.1cm}
The widely adopted ModelNet40~\cite{modelnet40} contains 9,843 training and 2,468 test 3D point clouds, which are sampled from the synthetic CAD models of 40 categories. We report the classification accuracy of existing methods before and after the voting in Table~\ref{t3}. As shown, our I2P-MAE achieves leading performance for both settings with only 1k input point number. For linear SVM without any fine-tuning, I2P-MAE-svm can already attain 93.4\% accuracy and exceed most previous works, indicating the powerful transfer ability. By fine-tuning the entire network, the accuracy can be further boosted by +.0.3\% accuracy, and achieve 94.1\% after the offline voting.
\vspace{-0.3cm}

\paragraph{Part Segmentation.}
The synthetic ShapeNetPart~\cite{shapenetpart} is selected from ShapeNet~\cite{chang2015shapenet} with 16 object categories and 50 part categories, which contains 14,007 and 2,874 samples for training and validation. We utilize the same segmentation head after the pre-trained encoder as previous works~\cite{zhang2022point,pang2022masked} for fair comparison. The head only conducts simple upsampling for point tokens at different stages and concatenates them alone the feature dimension as the output. Two types of mean IoU scores, mIoU$_C$ and mIoU$_I$ are reported in Table~\ref{t4}. For such highly saturated benchmark, I2P-MAE can still exert leading performance guided by the well learned 2D knowledge, e.g., +0.29\% and +0.25\% higher than Point-M2AE concerning the two metrics. This demonstrates that the 2D guidance also benefits the understanding for fine-grained point-wise 3D patterns.

\begin{table}[t!]
% \vspace{0.2cm}
\small
\centering
\begin{adjustbox}{width=0.92\linewidth}
	\begin{tabular}{lccc}
	\toprule
		\makecell*[l]{Method} &\ \ \ mIoU$_C$\ \ \ &\ \ \ mIoU$_I$\ \ \ \\
		\cmidrule(lr){1-1} \cmidrule(lr){2-2} \cmidrule(lr){3-3}
 		PointNet~\cite{qi2017pointnet}  &80.39 &83.70 \\
	    PointNet++~\cite{qi2017pointnet++} &81.85 &85.10 \\
	    DGCNN~\cite{dgcnn} &82.33 &85.20 \\
            PointMLP~\cite{pointmlp} &84.60 &86.10 \\
	    \cmidrule(lr){1-3}
	   %\cmidrule(lr){1-1} \cmidrule(lr){2-3} 
	    Transformer~\cite{pointbert} &83.42 &85.10 \\
	    {[P]} Transformer + OcCo~\cite{pointbert} &83.42 &85.10 \\
	    {[P]} Point-BERT~\cite{pointbert} &84.11 &85.60 \\
	    {[P]} MaskPoint~\cite{liu2022masked} &84.40 &86.00 \\
	    {[P]} Point-MAE~\cite{pang2022masked} &- &86.10 \\
	    {[P]} Point-M2AE~\cite{zhang2022point}\vspace{0.1cm} &84.86 &86.51 \\
	    \rowcolor{pink!12} \textbf{[P] I2P-MAE}\vspace{0.1cm} &\textbf{85.15} &\textbf{86.76}\\
	  \bottomrule
	\end{tabular}
\end{adjustbox}
\caption{\textbf{Part Segmentation on ShapeNetPart~\cite{shapenetpart}}. We report the average IoU scores (\%) for part categories and instances, respectively denoted as `mIoU$_C$' and `mIoU$_I$'.}
\label{t4}
% \vspace{-0.1cm}
\end{table}

% \paragraph{Few-shot Classification.}

\subsection{Ablation Study}
\label{s4.3}

In this section, we explore the effectiveness of different components in I2P-MAE. We utilize our final solution as the baseline for ablation and report the linear SVM classification accuracy (\%) for comparison by default.

\begin{table}[t!]
% \vspace{0.2cm}
\small
\centering
\begin{adjustbox}{width=\linewidth}
	\begin{tabular}{ccccc}
	\toprule
		\makecell*[c]{2D-Guided} &Visible &Ratio &ModelNet40 &OBJ-BG \\
		\cmidrule(lr){1-3} \cmidrule(lr){4-4} \cmidrule(lr){5-5} 
    \rowcolor{pink!12}\bf \checkmark  &\bf Important &\bf 0.8 &\bf 93.4 &\bf 87.1\vspace{0.05cm}\\
    -  &Random &0.8 &93.0 &86.2 \\
    \checkmark  &Unimportant &0.8 &92.8 &84.5 \\
    \checkmark  &Important &0.7 &92.5 &83.8 \\
    \checkmark  &Important &0.9 &92.8 &85.6 \\
	  \bottomrule
	\end{tabular}
\end{adjustbox}
\caption{\textbf{2D-guided Masking}. `Visible' denotes whether the preserved visible point tokens are more semantically important than the masked ones. `Ratio' denotes the masking ratio.}
\label{t5}
% \vspace{-0.1cm}
\end{table}

\begin{table}[t!]
\vspace{0.1cm}
\small
\centering
\begin{adjustbox}{width=\linewidth}
	\begin{tabular}{cccccc}
	\toprule
		\multicolumn{2}{c}{Tokens} &\multicolumn{2}{c}{Targets}  &\multirow{2}*{ModelNet40} &\multirow{2}*{OBJ-BG} \\
        \cmidrule(lr){1-2} \cmidrule(lr){3-4}Masked\hspace{-0.2cm} &Visible &\ \ \ 3D\ \ \ \  &\ \ 2D\ \ \ \\
		\cmidrule(lr){1-4} \cmidrule(lr){5-5} \cmidrule(lr){6-6} 
    \rowcolor{pink!12} $\checkmark$ &$\checkmark$ &M  &V &\bf 93.4 &\bf 87.1\vspace{0.05cm}\\
    \checkmark &- &M  &- &92.9 &84.7 \\
    - &\checkmark &- &V &91.9 &80.2 \\
    \checkmark &- &-  &M &91.2 &77.8 \\
    \checkmark &- &M  &M &92.6 &84.9 \\   
	  \bottomrule
	\end{tabular}
\end{adjustbox}
\caption{\textbf{2D-semantic Reconstruction}. We utilize `M' and `V' to denote the reconstruction targets of masked and visible tokens. `3D' and `2D' denote 3D coordinates and 2D semantics.}
\label{t6}
% \vspace{-0.1cm}
\end{table}

\paragraph{2D-guided Masking.}
In Table~\ref{t5}, we experiment different masking strategies for the masked autoencoding of I2P-MAE. The first row represents our I2P-MAE with 2D-guided masking, which preserves more semantically important tokens to be visible for the encoder. Compared to the second row with random masking, the guidance of 2D saliency maps contributes to +0.4\% and +0.9\% classification accuracy respectively on the two downstream datasets. Then, we reverse the token scores in the spatial semantic cloud, and instead mask the most important tokens. As shown in the third row, the SVM results are largely harmed by -0.9\% and -3.3\%, demonstrating the significance of encoding critical 3D structures in the encoder. Finally, we modify the masking ratio by $\pm$ 0.1, which controls the proportion between visible and masked tokens. The performance decay indicates that, the 2D-semantic and 3D-coordinate reconstruction are required to be well balanced for a properly challenging pre-text task. 

\begin{figure}[t]
  \centering
\vspace{0.2cm}
\includegraphics[width=0.45\textwidth]{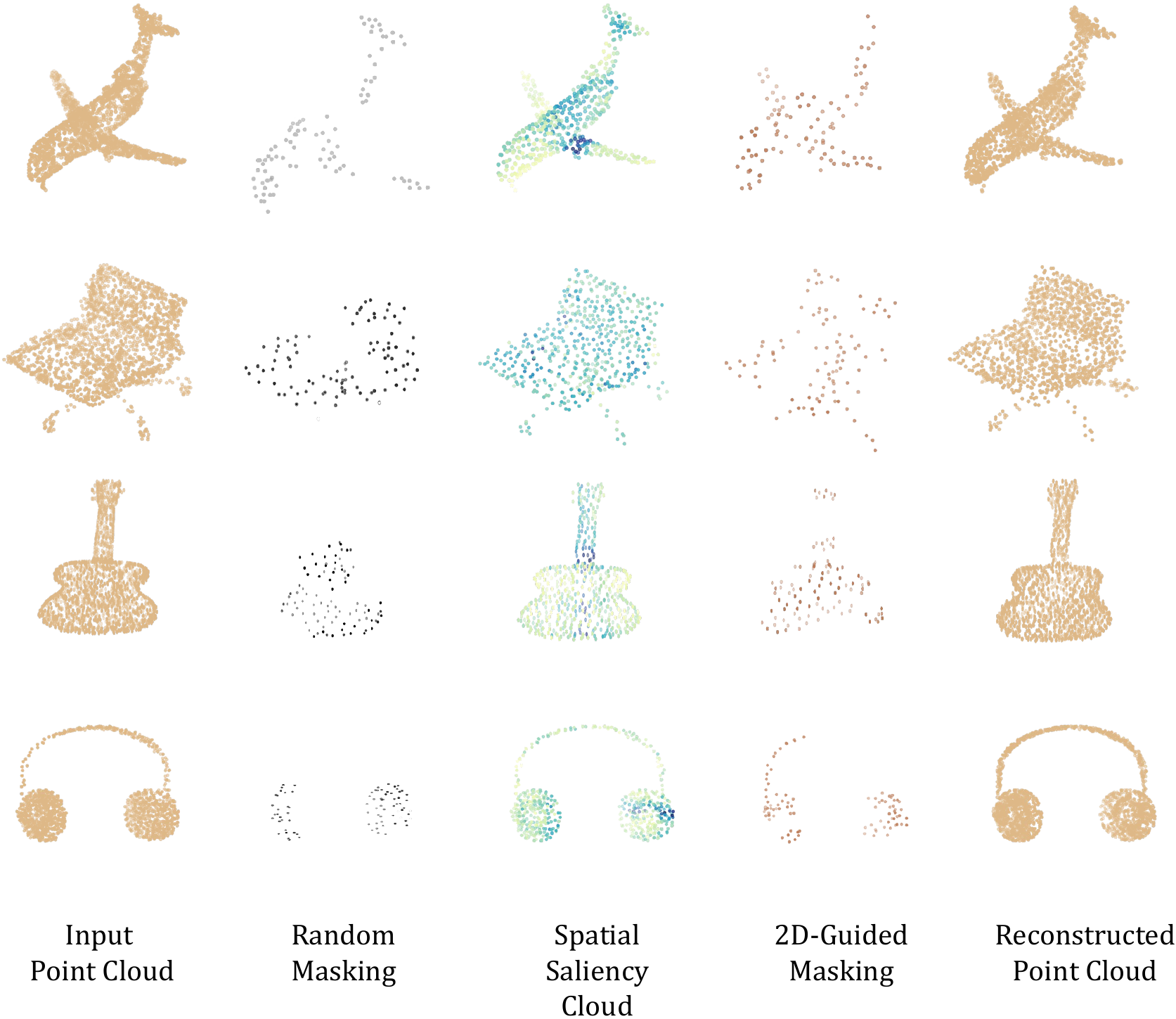}
\vspace{0.2cm}
   \caption{\textbf{Visualization of I2P-MAE.} Guided by the spatial saliency cloud, I2P-MAE's masking (the 4$^{th}$ column) preserves more semantically important 3D structures than random masking.}
    \label{fig7}
\vspace{0.1cm}
\end{figure}

\paragraph{2D-semantic Reconstruction.}
In Table~\ref{t6}, we investigate which groups of tokens are the best for learning 2D-semantic targets. The comparison of the first two rows reveals the effectiveness of reconstructing 2D semantics from visible tokens for point cloud pre-training, i.e., +0.5\% and +2.4\% classification accuracy. By only using 2D targets for either the visible or masked tokens (the 3$^{\text{rd}}$ and 4$^{\text{th}}$ rows), we verify that the 3D-coordinate reconstruction still plays an important role in I2P-MAE, which learns low-level geometric 3D patterns and provides complementary knowledge to the high-level 2D semantics. However, if the 3D and 2D targets are both reconstructed from the masked tokens (the last row), the network is restricted to learn 2D knowledge from the nonsignificant masked 3D geometries, other than the more discriminative parts. Also, assigning two targets on the same tokens might cause 2D-3D semantic conflicts. Therefore, the best-performing configuration is to reconstruct 2D and 3D targets separately from the visible and masked point tokens (the first row).

\begin{figure*}[t]
  \centering
% \vspace{0.2cm}
\includegraphics[width=0.7\textwidth]{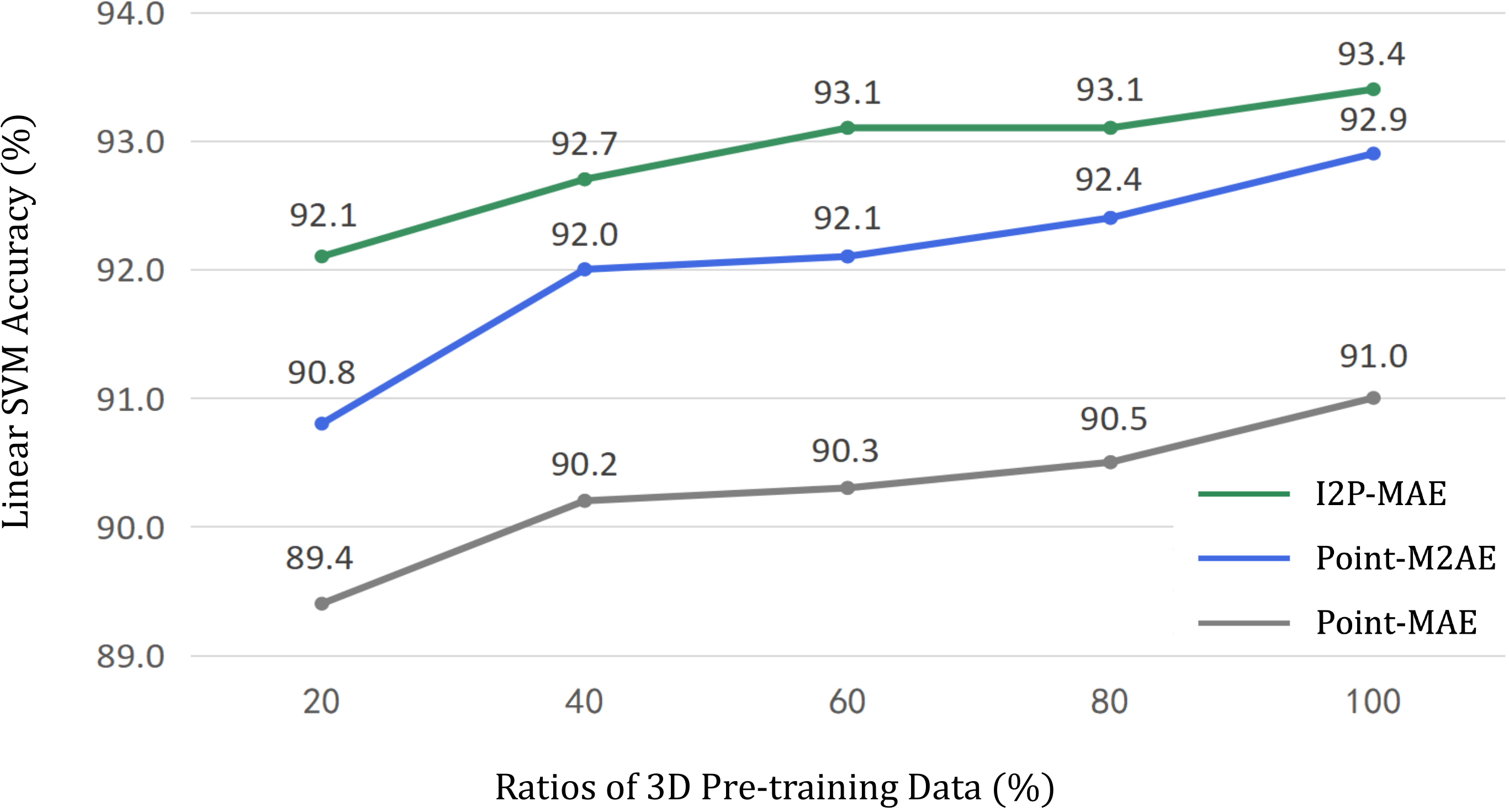}
\vspace{0.1cm}
   \caption{\textbf{Pre-training with Limited 3D Data.} We randomly sample different ratios of 3D data in ShapeNet~\cite{chang2015shapenet} for pre-training and report the linear SVM classification accuracy on ModelNet40~\cite{modelnet40}.
   Guided by 2D pre-trained models, I2P-MAE can acheive comparable performance to Point-M2AE~\cite{zhang2022point} with only half of the 3D data.}
    \label{supp_fig3}
\vspace{0.2cm}
\end{figure*}

\paragraph{Pre-training with Limited 3D Data.}
In Table~\ref{supp_t1} and Figure~\ref{supp_fig3}, we randomly sample the pre-training dataset, ShapeNet~\cite{chang2015shapenet}, by different ratios, and evaluate the performance of I2P-MAE when 3D data is further deficient. Aided by 2D pre-trained models, I2P-MAE still achieves competitive downstream accuracy in low-data regimes, especially for 20\% and 60\%, which outperforms Point-M2AE~\cite{zhang2022point} by +1.3\% and +1.0, respectively. \textbf{Importantly, with only 60\% of the pre-training, I2P-MAE (93.1\%) outperforms Point-MAE (91.0\%) and Point-M2AE (92.9\%) with full training data.} This indicates that our image-to-point learning scheme can effectively alleviate the need for large-scale 3D training datasets.

\begin{table}[t!]
\small
\centering
\begin{adjustbox}{width=\linewidth}
	\begin{tabular}{lcccccc}
	\toprule
	Method	&\makecell*[c]{\ 20\%\ } &\makecell*[c]{\ 40\%\ } &\makecell*[c]{\ 60\%\ } &\makecell*[c]{\ 80\%\ } &\makecell*[c]{\ 100\%\ } \\
		\cmidrule(lr){1-1} \cmidrule(lr){2-6}
  Point-MAE~\cite{pang2022masked} &89.4 &90.2  &90.3 &90.5 &91.0 \\
  Point-M2AE~\cite{zhang2022point}\vspace{0.05cm} &\underline{90.8} &\underline{92.0}  &\underline{92.1} &\underline{92.4} &\underline{92.9} \\  
    \rowcolor{pink!12} I2P-MAE &\bf 92.1 &\bf 92.7  &\bf93.1 &\bf93.1 &\bf 93.4\vspace{0.05cm}\\
    \textit{Improvement}  &\textcolor{blue}{+1.3} &\textcolor{blue}{+0.7} &\textcolor{blue}{+1.0} &\textcolor{blue}{+0.7} &\textcolor{blue}{+0.5}\\
	  \bottomrule
	\end{tabular}
\end{adjustbox}
\caption{\textbf{Pre-training with Limited 3D Data}. We conduct self-supervised 3D pre-training with different ratios of ShapeNet~\cite{chang2015shapenet}, and report the linear SVM accuracy (\%) on ModelNet40~\cite{modelnet40}. We highlight I2P-MAE's improvement over Point-M2AE~\cite{zhang2022point} in blue.}
\label{supp_t1}
% \vspace{-0.1cm}
\end{table}

\paragraph{Effectiveness of Pre-training.}
In Table~\ref{supp_t0}, we compare the performance on different downstream tasks between training from scratch and fine-tuning after pre-training. For ScanObjectNN~\cite{scanobjectnn}, the pure 3D pre-training without image-to-point learning (`w/o 2D Guidance') can improve the classification accuracy by +1.09\%, and our proposed 2D-to-3D knowledge transfer further boosts the performance by +2.59\%. Similar improvement can be observed on other downstream datasets, which demonstrates the significance of the pre-training of I2P-MAE. 
% In Figure~\ref{supp_fig1} and~\ref{supp_fig2}, we show the comparison of learning curves on two shape classification datasets. Our image-to-point pre-training can largely accelerate the convergence speed during downstream fine-tuning.

\section{Visualization}
To ease the understanding of our approach, we visualize the input point cloud, random masking, spatial saliency cloud, 2D-guided masking, and the reconstructed 3D coordinates in Figure~\ref{fig7}. Guided by the semantic scores from 2D pre-trained models (darker points indicate higher scores), the masked point cloud largely preserves the significant parts of the 3D shape, e.g., the main body of an airplane, the grip of a guitar, the frame of a chair and headphone.
In this way, the 3D network can learn more discriminative features by reconstructing these visible 2D semantics.

\begin{table}[t!]
\small
\centering
% \vspace{0.1cm}
\begin{adjustbox}{width=0.96\linewidth}
	\begin{tabular}{lcccc}
	\toprule
        Dataset &\makecell*[c]{From\\Scratch}  &\makecell*[c]{w/o 2D\\Guidance} &\makecell*[c]{\textbf{I2P-MAE}}\\
		\cmidrule(lr){1-1} \cmidrule(lr){2-4}
  ScanObjectNN~\cite{scanobjectnn} &86.34 &87.52 &\bf90.11 \\  
  ModelNet40~\cite{modelnet40} &92.46 &93.43 &\bf93.72 \\
  ShapeNetPart~\cite{shapenetpart}&86.38 &86.51 &\bf86.76 \\
  ModelNet40-FS~\cite{modelnet40} &91.20 &95.00 &\bf95.50 \\
	  \bottomrule
	\end{tabular}
\end{adjustbox}
\caption{\textbf{Effectiveness of Pre-training.} We report the downstream performance (\%) of training from scratch and fine-tuning after pre-training. `w/o 2D Guidance' denotes the pre-training without learning from 2D pre-trained models. We adopt the PB-T50-RS split of ScanObjectNN and denote the 10-way 20-shot split for few-shot classification as ModeNet-FS.}
\label{supp_t0}
% \vspace{-0.05cm}
\end{table}

\section{Conclusion}
In this paper, we propose I2P-MAE, a masked point modeling framework with effective image-to-point learning schemes. We introduce two approaches to transfer the well learned 2D knowledge into 3D domains: 2D-guided masking and 2D-semantic reconstruction. Aided by the 2D guidance, I2P-MAE learns superior 3D representations and achieves state-of-the-art performance on 3D downstream tasks, which alleviates the demand for large-scale 3D datasets. For future work, not limited to masking and reconstruction, we will explore more sufficient image-to-point learning for 3D masked autoencoders, e.g., point token sampling and 2D-3D class-token contrast. Also, we expect our pre-trained models to benefit wider ranges of 3D tasks, e.g., 3D object detection and visual grounding.

\begin{figure}[t]
  \centering
% \vspace{0.2cm}
\includegraphics[width=0.47\textwidth]{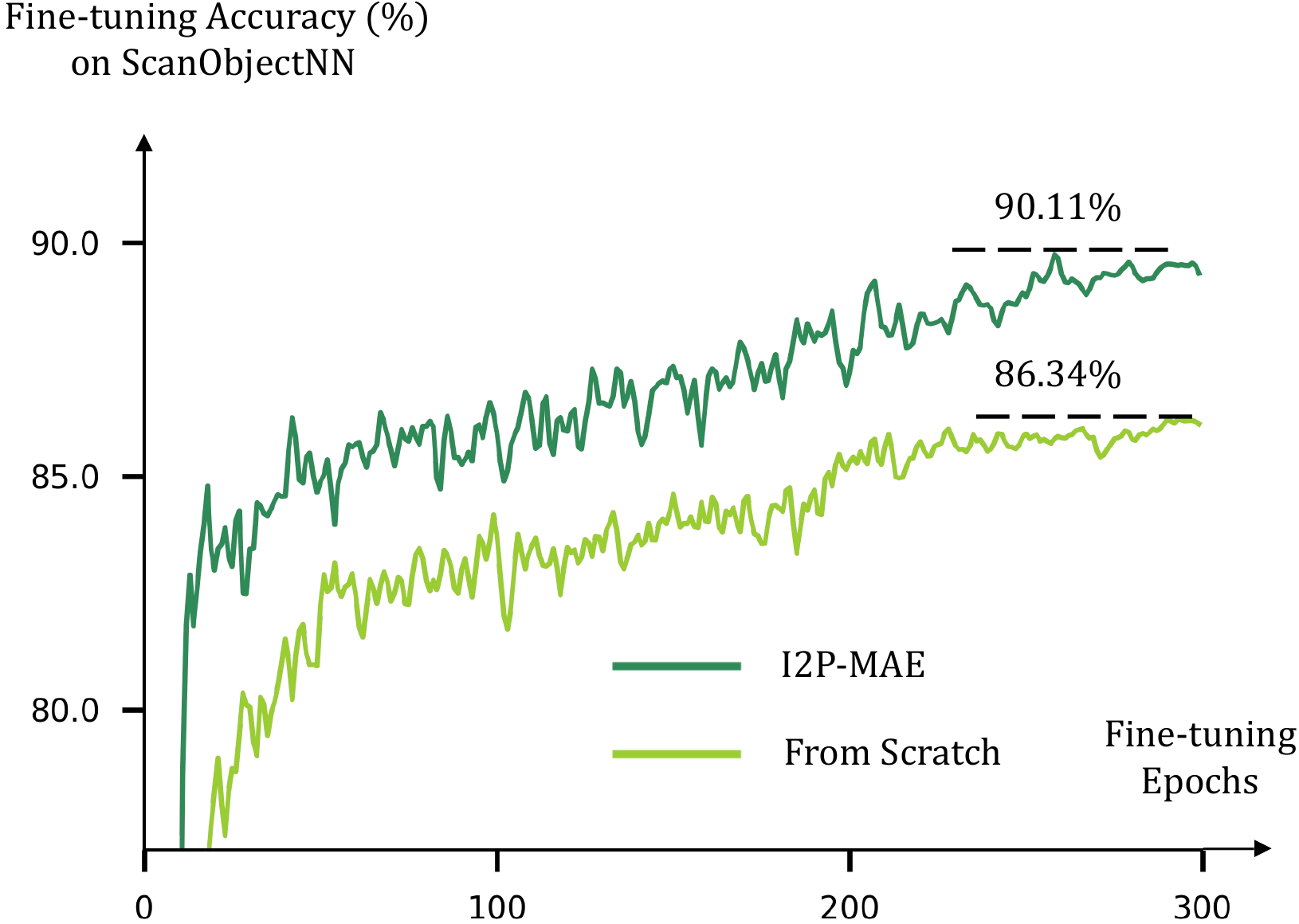}
\vspace{0.1cm}
   \caption{\textbf{I2P-MAE Fine-tuning vs. Training from Scratch on SacnObjectNN~\cite{scanobjectnn}.} We report the fine-tuning accuracy on the PB-T50-RS split of ScanObjectNN.}
    \label{supp_fig1}
% \vspace{-0.2cm}
\end{figure}

\section{Appendix}
\vspace{0.3cm}

\subsection{Implementation Details}

In this section, we present the detailed model configuration and training settings for fine-tuning I2P-MAE on downstream tasks. All experiments are conducted on a single RTX 3090 GPU.

\paragraph{Shape Classification.}
For both ModelNet40~\cite{modelnet40} and ScanObjectNN~\cite{scanobjectnn}, we fine-tune I2P-MAE for 300 epochs with a batch size 32. We adopt AdamW~\cite{kingma2014adam} optimizer with a learning rate 0.0005 and weight decay 0.05, and utilize cosine scheduler with a 10-epoch warm-up. We append a 3-layer MLP after I2P-MAE's encoder as the classification head. For ScanObjectNN, we
adopt max and average pooling to respectively summarize the point tokens from the encoder, and concatenate the two global features along the feature dimension for the head. I2P-MAE follows existing methods~\cite{pang2022masked,zhang2022point} to take 2,048 points as input, and adopts random scaling with rotation as data augmentation. For ModelNet40, we element-wisely add the two global features for the classification head. I2P-MAE takes 1,024 points as input, and adopts random scaling with translation as data augmentation.

\paragraph{Part Segmentation.}
On ShapeNetPart~\cite{shapenetpart}, we fine-tune I2P-MAE for 300 epochs with a batch size 16. We also adopt AdamW~\cite{kingma2014adam} optimizer with a learning rate 0.0002 and weight decay 0.00005, and utilize cosine scheduler with a 10-epoch warm-up. For fair comparison, we experiment with the same segmentation head and training settings as Point-M2AE~\cite{zhang2022point}.

\subsection{Additional Ablation Study}
\vspace{0.2cm}

\begin{figure}[t]
  \centering
% \vspace{0.2cm}
\includegraphics[width=0.44\textwidth]{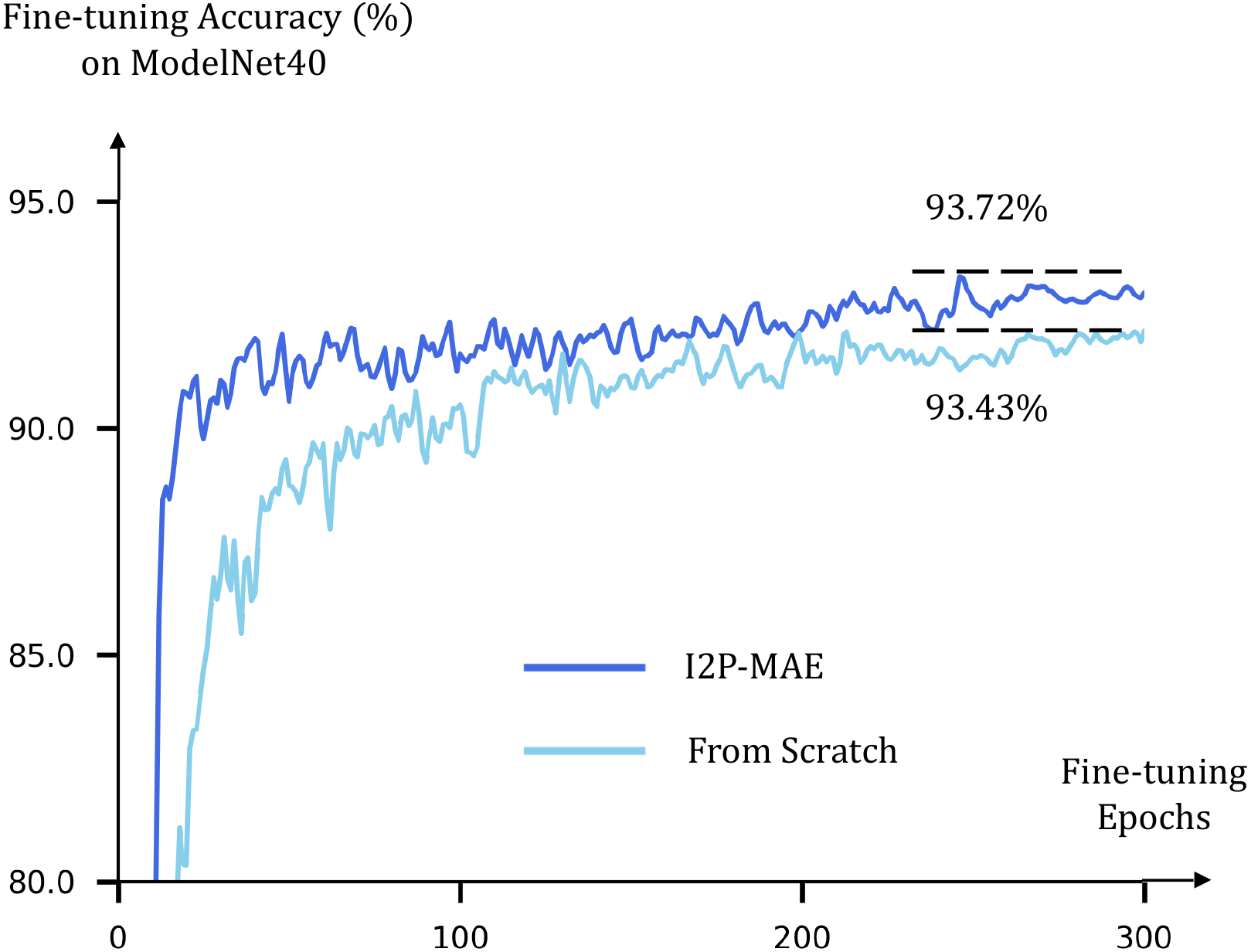}
\vspace{0.1cm}
   \caption{\textbf{I2P-MAE Fine-tuning vs. Training from Scratch on ModelNet40~\cite{modelnet40}.} We report the fine-tuning accuracy on ModelNet40.}
    \label{supp_fig2}
% \vspace{-0.2cm}
\end{figure}

\paragraph{Learning Curves of Pre-training.}
In Figure~\ref{supp_fig1} and~\ref{supp_fig2}, we show the comparison of training I2P-MAE from scratch and fine-tuning after pre-training on two shape classification datasets. Our image-to-point pre-training can largely accelerate the convergence speed during fine-tuning and the final classification accuracy, indicating the effectiveness of the 2D-to-3D knowledge transfer.

\paragraph{Projected View Number.}
In Table~\ref{supp_t2} (1$^{st}$ and 2$^{nd}$ rows), we show how the number of projected views affect the performance of I2P-MAE. As default, we project the point cloud into 3 views along the $x, y, z$ axes. For the view number 1 and 2, we enumerate all possible projected views along $x, y, z$ axes, and report the highest results in the table. As shown, using less views would harm the pre-training performance, which constrains 2D pre-trained models from `seeing' complete 3D shapes due to occlusion. Instead, the 3D network can learn more comprehensive high-level semantics from the 2D representations of all three views.

\paragraph{3D Sailency Cloud and 2D-semantic Target.}
We first investigate how to aggregate multi-view 2D sailency maps as the sailency cloud for 2D-guided masking in Table~\ref{supp_t2} (3$^{rd}$ and 4$^{th}$ rows). Compared to assigning the maximum or minimum score to a certain point, averaging the 2D sailency scores from different views achieves the best performance. Then, we explore how to generate the 2D-semantic targets from multi-view 2D features in Table~\ref{supp_t2} (5$^{th}$ row). The results indicate that, concatenating 2D features between different views performs better than averaging them, which preserves more diverse 2D semantics for reconstruction.

\begin{table*}[t!]
\small
\centering
\begin{adjustbox}{width=0.68\linewidth}
	\begin{tabular}{lc c c c c}
	\toprule
% 	\multicolumn{5}{c}{Few-shot Classification on ModelNet40~\cite{modelnet40}} \\
% 	\midrule
		\makecell*[c]{\multirow{2}*{Method}} &\multicolumn{2}{c}{5-way} &\multicolumn{2}{c}{10-way}\\
		 \cmidrule(lr){2-3} \cmidrule(lr){4-5}
		 &10-shot &20-shot &10-shot &20-shot\\
		 \cmidrule(lr){1-1} \cmidrule(lr){2-5}
		DGCNN~\cite{dgcnn} &91.8\ $\pm$\ 3.7 &93.4\ $\pm$\ 3.2 &86.3\ $\pm$\ 6.2 &90.9\ $\pm$\ 5.1\\
		{[P]} DGCNN + OcCo~\cite{occo} &91.9\ $\pm$\ 3.3 &93.9\ $\pm$\ 3.1 &86.4\ $\pm$\ 5.4 &91.3\ $\pm$\ 4.6\\
		\cmidrule(lr){1-5}
	    Transformer~\cite{pointbert} &87.8\ $\pm$\ 5.2 &93.3\ $\pm$\ 4.3 &84.6\ $\pm$\ 5.5 &89.4\ $\pm$\ 6.3\\
		{[P]} Transformer + OcCo~\cite{pointbert} &94.0\ $\pm$\ 3.6 &95.9\ $\pm$\ 2.3 &89.4\ $\pm$\ 5.1 &92.4\ $\pm$\ 4.6\\
		{[P]} Point-BERT~\cite{pointbert} &94.6\ $\pm$\ 3.1 &96.3\ $\pm$\ 2.7 &91.0\ $\pm$\ 5.4 &92.7\ $\pm$\ 5.1\\
        {[P]} MaskPoint~\cite{liu2022masked} &95.0\ $\pm$\ 3.7 &97.2\ $\pm$\ 1.7 &91.4\ $\pm$\ 4.0 &93.4\ $\pm$\ 3.5\\
        {[P]} Point-MAE~\cite{pang2022masked} &96.3\ $\pm$\ 2.5 &97.8\ $\pm$\ 1.8 &92.6\ $\pm$\ 4.1 &95.0\ $\pm$\ 3.0\\   
        {[P] Point-M2AE~\cite{zhang2022point}} &96.8\ $\pm$\ 1.8& 98.3\ $\pm$\ 1.4 &92.3\ $\pm$\ 4.5 &95.0\ $\pm$\ 3.0\vspace{0.1cm}\\
	   \rowcolor{pink!12} \textbf{[P] I2P-MAE} &\textbf{97.0\ $\pm$\ 1.8}& \textbf{98.3\ $\pm$\ 1.3}&\textbf{92.6\ $\pm$\ 5.0} &\textbf{95.5\ $\pm$\ 3.0}\vspace{0.1cm}\\
	\bottomrule
	\end{tabular}
\end{adjustbox}
\caption{\textbf{Few-shot Classification on ModelNet40~\cite{modelnet40}}. We report the average classification accuracy ($\%$) with the standard deviation ($\%$) of 10 independent experiments. [P] denotes to fine-tune the models after self-supervised pre-training.
}
 \label{supp_t5}
% \vspace*{0.2cm}
\end{table*}

\begin{table}[t!]
\small
\centering
\begin{adjustbox}{width=0.97\linewidth}
	\begin{tabular}{cccc}
	\toprule
	\makecell*[c]{Projected\\Views}	&\makecell*[c]{3D Sailency\\Cloud} &\makecell*[c]{2D-semantic\\Target} &\makecell*[c]{ModelNet40} \\
		\cmidrule(lr){1-3} \cmidrule(lr){4-4}
    2 &Ave &Concat &93.0\\
    1 &Ave &Concat &92.9\\
    % \cmidrule(lr){1-4}
    \cmidrule(lr){1-3} \cmidrule(lr){4-4}
    3 &Max &Concat &93.3\\
    3 &Min &Concat &92.8\\
    % \cmidrule(lr){1-4}
    \cmidrule(lr){1-3} \cmidrule(lr){4-4}
    3\vspace{0.05cm} &Ave &Ave &93.1\\
    \rowcolor{pink!12} 3 &Ave &Concat &\bf 93.4\vspace{0.05cm}\\
	  \bottomrule
	\end{tabular}
\end{adjustbox}
\caption{\textbf{Different Image-to-Point Settings.} `Ave', `Max', `Min', `Concat' denote different operations to aggregate multi-view 2D representations. We report the linear SVM accuracy (\%).}
\label{supp_t2}
% \vspace{-0.1cm}
\end{table}

\paragraph{Fine-tuning Settings.}
In Table~\ref{supp_t4}, we experiment different fine-tuning settings for downstream shape classification on the two datasets. For the point tokens from the encoder, `Max Only' and `Ave Only' denote applying either max or average pooling to summarize global features for the classification head. `Add' or `Concat' denotes to add or concatenate the two global features after max and average pooling. We observe that, `Add' and `Concat' perform the best for ModelNet40~\cite{modelnet40} and ScanObjectNN~\cite{scanobjectnn}, respectively.

\vspace{0.1cm}
\subsection{Few-shot Classification}

We fine-tune I2P-MAE for few-shot classification on ModelNet40~\cite{modelnet40} in Table~\ref{supp_t5}. Following previous work~\cite{pointbert,zhang2022point,pang2022masked}, we adopt the same training settings and few-shot dataset splits, i.e., 5-way 10-shot, 5-way 20-shot, 10-way 10-shot, and 10-way 20-shot. With limited downstream fine-tuning data, our I2P-MAE exhibits competitive performance among existing methods, e.g., +0.5\% classification accuracy to Point-M2AE~\cite{zhang2022point} on the 10-way 20-shot split.

\begin{table}[t!]
\small
\centering
\begin{adjustbox}{width=0.96\linewidth}
	\begin{tabular}{ccc}
	\toprule
        Settings &\makecell*[c]{ModelNet40~\cite{modelnet40}}  &\makecell*[c]{ScanObjectNN~\cite{scanobjectnn}} \\
		\cmidrule(lr){1-1} \cmidrule(lr){2-2} \cmidrule(lr){3-3}
  
  Max Only &93.31 &89.90 \\
  Ave Only &93.56 &88.83 \\
  Add &\bf93.72 &89.20 \\
  Concat &93.23 &\bf90.11 \\
  % Class Token & & \\
	  \bottomrule
	\end{tabular}
\end{adjustbox}
\caption{\textbf{Fine-tuning Settings.} We experiment different approaches to summarize global features of the encoder for downstream fine-tuning. We report the fine-tuning accuracy (\%) on ModelNet40 and the PB-T50-RS split of ScanObjectNN.}
\label{supp_t4}
% \vspace{-0.1cm}
\end{table}

\subsection{Additional Visualization}

\begin{figure*}[t]
  \centering
% \vspace{0.2cm}
\includegraphics[width=\textwidth]{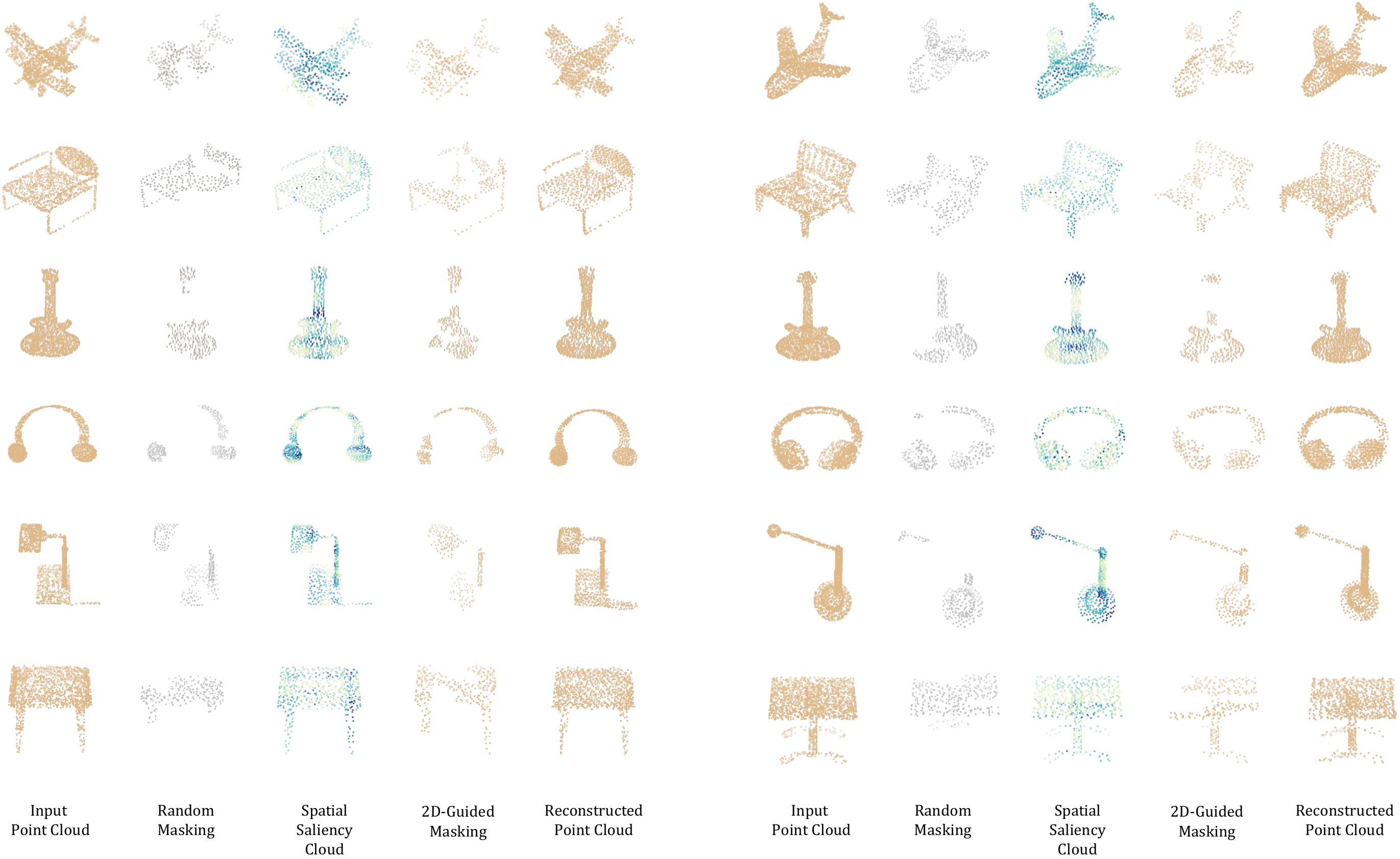}
\vspace{0.1cm}
   \caption{\textbf{Additional Visualization of I2P-MAE.} Guided by the spatial saliency cloud, I2P-MAE's masking (the 4$^{th}$ and 9$^{th}$ columns) preserves more semantically important 3D structures than random masking.}
    \label{supp_fig4}
% \vspace{-0.2cm}
\end{figure*}

In Figure~\ref{supp_fig4}, we additionally visualize the input point cloud, random masking, spatial saliency cloud, 2D-guided masking, and the reconstructed 3D coordinates, respectively. 
As shown, the 2D-guided masking can preserve the semantically important 3D geometries guided by the spatial sailency cloud (darker points indicate higher scores).
In this way, I2P-MAE can inherit more significant 2D knowledge through the 2D-semantic reconstruction of the unmasked visible parts.

% \clearpage

%%%%%%%%% REFERENCES
{\small
\bibliographystyle{ieee_fullname}
\bibliography{egbib}
}

\end{document}